\documentclass[letterpaper, 10 pt, conference]{ieeeconf} 
\IEEEoverridecommandlockouts

\usepackage{graphicx}
\usepackage{siunitx}
\usepackage[ruled,vlined]{algorithm2e}

\usepackage{amsmath} 

\usepackage{subcaption}
\usepackage{paralist} 
\usepackage{xcolor}
\usepackage{xurl}
\usepackage{booktabs} 
\usepackage{multirow}
\usepackage{amsmath,amssymb,amsfonts}
\usepackage{mathtools}
\usepackage{stackengine}
\usepackage{flushend} 

\usepackage{esvect}
\usepackage{hyperref}
\hypersetup{
    colorlinks=true,
    linkcolor=blue,
    filecolor=magenta,      
    urlcolor=cyan,
}

\title{\LARGE \bf

Monolithic vs. hybrid controller for multi-objective Sim-to-Real learning}

\author{Atakan Dag$^{1}$, Alexandre Angleraud$^{2}$, Wenyan Yang$^{1}$, Nataliya Strokina$^{1}$, Roel S. Pieters$^{2}$,\\ Minna Lanz$^{2}$, and Joni-Kristian K{\"a}m{\"a}r{\"a}inen$^{1}$

\thanks{$^{1}$Computing Sciences and $^{2}$Automation Technology and Mechanical Engineering, Tampere University, Finland.}
}

\begin{document}
\maketitle
\thispagestyle{empty}
\pagestyle{empty}

\begin{abstract}
Simulation to real (Sim-to-Real) is an attractive approach to construct controllers for robotic tasks that are easier to simulate than to analytically solve. Working Sim-to-Real solutions have been demonstrated for tasks with a clear single objective such as "reach the target". Real world applications, however, often consist of multiple simultaneous objectives such as "reach the target" but "avoid obstacles". A straightforward solution in the context of reinforcement learning (RL) is to combine multiple objectives into a multi-term reward function and train a single monolithic controller. Recently, a hybrid solution based on pre-trained single objective controllers and a switching rule between them was proposed. In this work, we compare these two approaches in the multi-objective setting of a robot manipulator to reach a target while avoiding an obstacle. Our findings show that the training of a hybrid controller is easier and obtains a better success-failure trade-off than a monolithic controller. The controllers trained in simulator were verified by
a real set-up.

\end{abstract}

\section{INTRODUCTION}

While recent advances in simulation-driven Reinforcement Learning
(RL) are impressive~\cite{lillicrap2015continuous,TheRLBook,luo2020accelerating}, the bar by which these are measured in robotics is their capability to be deployed to real robot tasks. In the "Sim-to-Real" approach~\cite{Rusu-2017-CoRL,Matas-2018-CoRL} the objective is to
train a controller in a simulated environment and then transfer the
controller to the real environment. Successful Sim-to-Real controllers have been demonstrated for object manipulation tasks~\cite{Wen-2018-robio,Sangiovanni-2018-ecc,Sangiovanni-2021-ieeecsl,Pham-2018-icra,Zhang-2019-smartcloud,Zhang-2020-rssw,luo2020accelerating}. These tasks have one main objective, but practical applications
often need multiple simultaneous objectives such as "reach the target" and "avoid obstacles". For example, in collaborative human-robot manufacturing
the robot should complete its task without compromising human safety~\cite{Halme-2018-cirp,Hietanen-2019-roman,Hietanen-RCIM-2020}.

A straightforward solution in the spirit of
reinforcement learning (RL) is to combine multiple
objectives into the reward function and train a
{\em monolithic} controller. This was demonstrated for
robot reaching and collision avoidance by Pham~et~al.~\cite{Pham-2018-icra} who trained a controller
in a simulator and then deployed it to a real robot.
Pham~et~al. defined a reward function that includes
positive reward for reaching the target and negative reward
(penalty) for being too close to an obstacle.
The design burden of the monolithic approach is in
"reward engineering" to find and tune a suitable reward function.
An alternative approach was recently proposed by
Sangiovanni~et~al.~\cite{Sangiovanni-2021-ieeecsl} who propose
a {\em hybrid} controller that consists of a number of
pre-trained controllers for each task and
a switching rule
(the closest obstacle being too close to the manipulator).
In their case the design burden is in defining a suitable
switching rule. Moreover, Sangiovanni~et~al. did not experiment
their hybrid controller with a real robot. In our work,
we adopt the hybrid controller model for
multi-objective Sim-to-Real tasks.

\begin{figure}[t]
    \centering
    \includegraphics[width=1.0\linewidth]{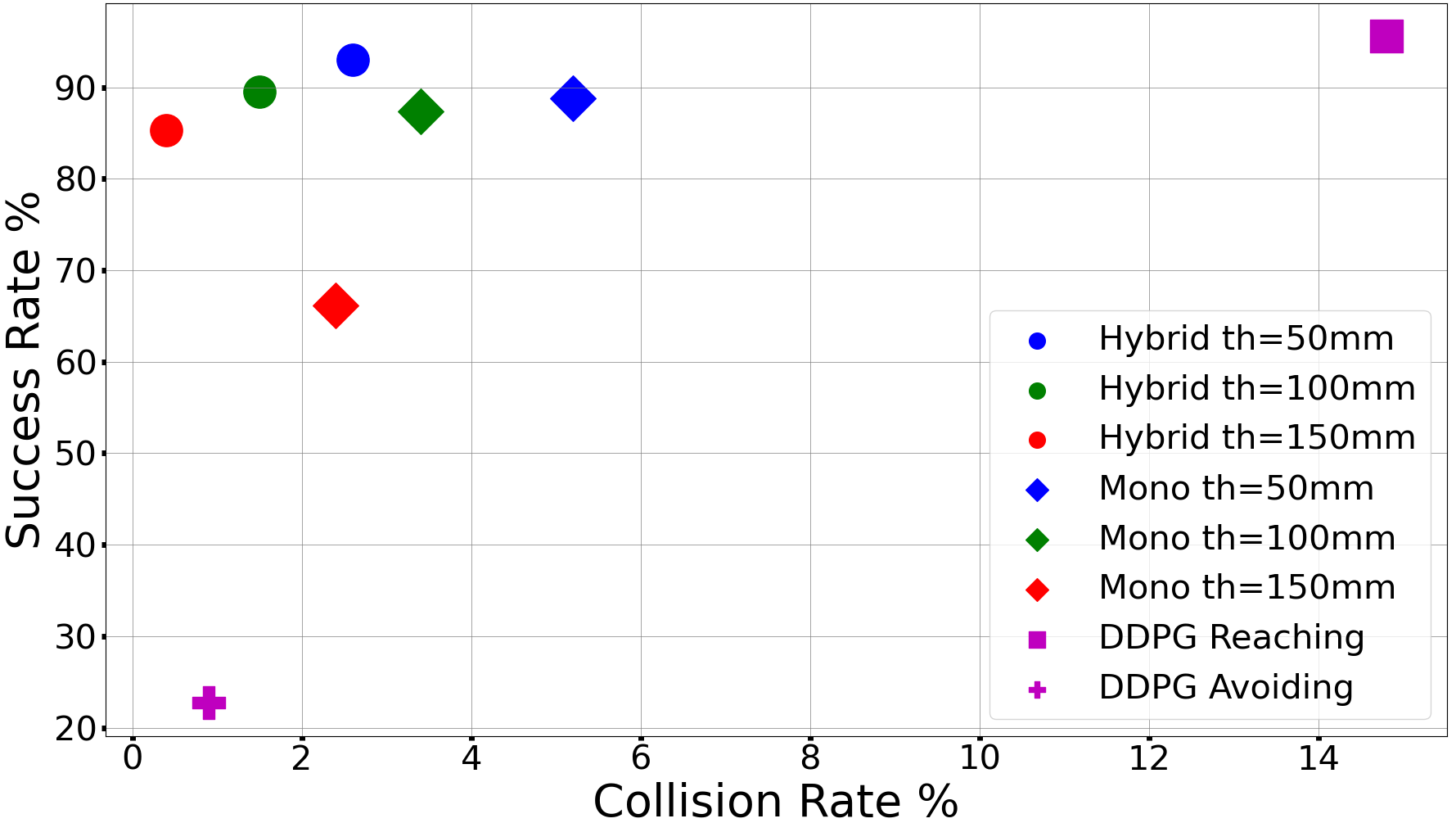}\\
    \vspace{\medskipamount}
    \includegraphics[width=1.0\linewidth]{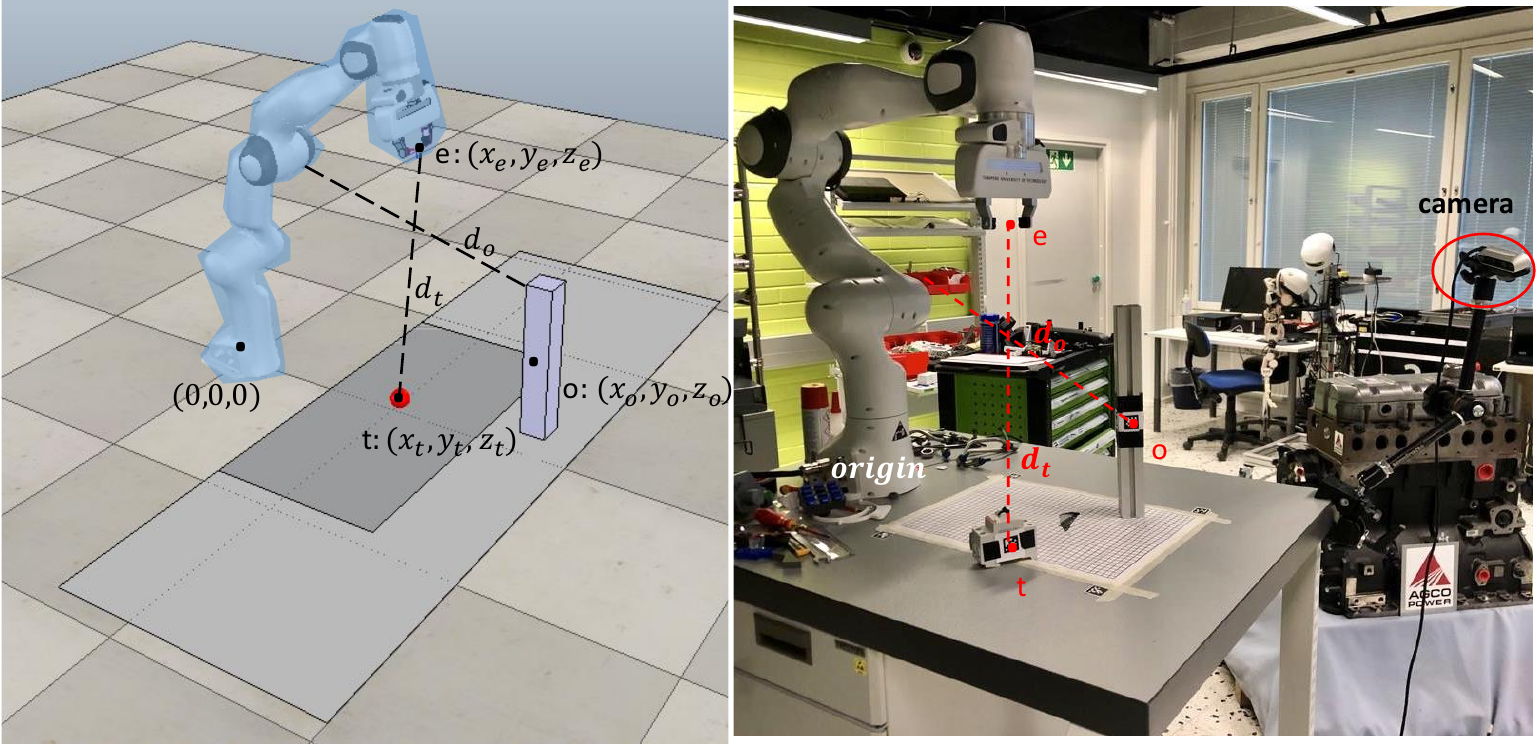}
    \caption{Simulation results for the multi-objective "Reach the target and avoid the obstacle" task (top graph). Average success and collision rates over 1000 episodes for the monolithic~\cite{Pham-2018-icra} and hybrid~\cite{Sangiovanni-2021-ieeecsl} controllers on three
    operation points each. The monolithic controller (diamonds) is not able to 
    achieve the superior success-failure (collision) rates of
    the hybrid controller (circles). Moreover, the monolithic controller
    needs to be re-trained for each new operation point while the hybrid controller is trained only once. The simulated success and collision rates are verified by a real setup (bottom images).}
    \label{fig:teaser}
\end{figure}

In this work, we compare two approaches, monolithic and hybrid,
in the multi-objective setting of a robot manipulator
to reach the target and avoid collision with an obstacle.
The problem is solved using the Sim-to-Real approach where all
controllers are trained in a simulator and then transferred
to the real environment. Our main contributions are:
\begin{enumerate}
\item reinforcement learning based implementations of the monolithic and hybrid controllers for the "reach and avoid" task in a simulated environment,
\item comparison results of the two controller models in the simulated environment indicating that the hybrid model is easier to train and tune, and it achieves better success-failure trade-off (Figure~\ref{fig:teaser}); and
\item verification of the results in Sim-to-Real by transferring and experimenting the controllers with a real Franka Emika Panda manipulator.
\end{enumerate}
All code and data are publicly available at \url{https://github.com/atakandag/multi-objective-sim2real}.

\section{RELATED WORK}

\noindent\textit{Sim-to-Real --} The Sim-to-Real research challenge is to
train a controller in a simulated environment for the target task and then
transfer the controller to a real environment. A popular benchmark task is
robot reaching where a static or moving target is reached in the presence
of one or multiple static or dynamic obstacles. Reinforcement learning based
controllers have been demonstrated in simulators~\cite{Wen-2018-robio,Sangiovanni-2018-ecc,Sangiovanni-2021-ieeecsl},
and a number of Sim-to-Real setups have been demonstrated~\cite{Pham-2018-icra,Zhang-2019-smartcloud,Zhang-2020-rssw,luo2020accelerating}. These works differ by the selected RL algorithm, sensor inputs and RL reward functions used.

Matas~et~al.~\cite{Matas-2018-CoRL} study the Sim-to-Real problem for deformable objects that are more difficult to model than the rigid objects. Golemo~et~al.~\cite{Golemo-2018-CoRL}
address a more general problem of how to
close the gap between the simulated worlds and the real world with more noise and distortions.
Golemo~et~al. propose a Neural-Augmented Simulation (NAS) method that provides better policies than those trained only on simulated data. Peng~et~al.~\cite{Peng-2018-icra} show that by adding randomness to system dynamics during training the performance in real world tasks is improved.
The most similar to our work are Pham~et~al.~\cite{Pham-2018-icra} and Sangiovanni~et~al.~\cite{Sangiovanni-2021-ieeecsl} who both address the problem of
a multi-objective task where the target must be reached (success) but without
hitting an obstacle (collision/failure). In our work, we compare the monolithic
controller used by Pham~et~al. and the hybrid controller of Sangiovanni~et~al. for the
multi-objective reaching task.

\vspace{\medskipamount}\noindent\textit{Multi-objective RL --} Most of the current multi-objective RL approaches are tested and validated in simulated toy environments and often with a discrete state-action space~\cite{Liu2015}. Single-policy methods suggest construction of a compound reward function that combines rewards for each objective with different
weights~\cite{Pham-2018-icra,Chen2019}.
There is a number of "multi-policy" methods that
assume a set of Pareto optimal policies to solve the task and the goal is to
approximate these policies~\cite{Natarajan2005}. A recent work in this direction  ~\cite{Xu2020} proposes an evolutionary learning algorithm to find Pareto set approximation
for continuous control problems. However, results have only been demonstrated for
tasks in simulated environments. The hybrid controller model of Sangiovanni et al. provides a new alternative which combines single
task controllers for a more complex multi-objective task.

\section{METHODS}\label{sec:methods}
Reinforcement Learning (RL) provides a model-free approach to
learn a controller from real or simulated episodes of a task. The
objective is to learn a policy controller that
maximizes the system reward at all system states. RL approaches
can be divided to tabular and approximate methods~\cite{TheRLBook}.
Tabular methods, such as Temporal Difference Learning~\cite{Sutton-1988-ml} and Q-learning~\cite{Watkins-1989-phd}, are suitable for problems with a small number of states and actions. Approximate methods are more suitable for large scale problems since
the value and policy functions are learned by
continuous function approximators such as neural networks. A
neuron-type function approximation was already proposed as {\em actor-critic}
by Barto~et~al.~\cite{Barto-1983-ieeescm},
but they decoded continuous inputs to a small number of
discrete states.

Continuous policy approximation was proposed for the
REINFORCE algorithm in~\cite{Williams-1992-ml}. A more general
definition for the {\em Policy Gradient} (PG) approach was presented in~\cite{Sutton-2000-nips}. Silver~et~al.~\cite{Silver-2014-icml}
introduced the Deterministic Policy Gradient (DPG) method that
optimizes policy function using an estimated action-value function
(actor-critic). Neural networks can be used as the function
approximators for DPG methods. For example, Deep DPG (DDPG)
was introduced by Lillicrap~et~al.~\cite{lillicrap2015continuous}
and it was extended with 
Hindsight Experience Replay in~\cite{andrychowicz2017hindsight}.
The advanced methods are particularly useful for sparse rewards,
but for simplicity we adopt the DDPG algorithm and
use dense continuous rewards.

\subsection{Deep Deterministic Policy Gradient (DDPG)}

Deep Deterministic Policy Gradient (DDPG) by Lillicrap~et~al.~\cite{lillicrap2015continuous} is a popular choice
for off-policy RL. DDPG adopts the Actor-Critic architecture~\cite{Barto-1983-ieeescm} where policy and value functions
are estimated separately. In DDPG, both the state-dependent actor ($\mu$) and the state-value critic ($Q$) functions are estimated by neural networks.
The actor takes the current state as input and outputs directly control
action values rather than probabilities of actions. The
critic takes the current state and predicted action as inputs
and outputs an estimate of the state-action value function $Q(s,a)$.
The critic provides a baseline which reduces variance of the gradient
based policy update steps. To further stabilize training, DDPG
uses target networks, $\mu'$ and $Q'$, in its update steps.
Target functions are just time delayed versions of the original functions. 
Since it is an off-policy algorithm, DDPG uses an experience replay which stores experiences $(s_t,a_t,r_t,s_{t+1})$: at every timestep t, an agent observes state $s_t$, conducts action $a_t$, gets reward $r_t$ for that action and observes the next state $s_{t+1}$. 
Actions predicted by the randomly initialized action network are obtained from

\begin{equation}
    a_t = \mu(s_t| \theta^\mu) + \mathcal{N}_t, 
\end{equation}

where $\theta^\mu$ refers to the weights of the actor network and $\mathcal{N}_t$ is the noise added for exploration. In each step, a random batch of experiences $(s_i,a_i,r_i,s_{i+1})$ is sampled from the experience replay buffer and the critic is updated by minimizing the MSE loss between the expected $Q$ value predicted by the critic network and the target $Q$ value using the following equations:
\begin{equation}
\begin{split}
&L_{critic} = MSE(y_t, Q(s_{t+1}, a_{t+1})) \\
&y_t = r_t + \gamma Q'\left(s_{t+1} , \mu'(s_{t+1})\right)
\end{split}\enspace .
\end{equation}
The main target is to find a policy that maximizes the expected return $\mathbb{E}[Q(s,a)\mid_{s=s_t,a_t=\mu(s_t)}]$. For that purpose, the actor is updated by the DPG policy gradient step~\cite{Silver-2014-icml}
\begin{equation}
    \nabla_{\theta^\mu} J  \approx
    \frac{1}{N}\sum\limits_{i}\nabla_a Q(s,a | \theta^Q) \mid_{s=s_i, a= \mu(s_i)} \nabla_{\theta^\mu}\mu(s | \theta^\mu)\mid_{s_i},
\end{equation}
where $N$ is the size of the batch. Updates are made to the target networks as the last step ($\theta^{Q'} \leftarrow \tau\theta^Q+(1-\tau)\theta^{Q'}$ and
$\theta^{\mu'} \leftarrow \tau\theta^\mu+(1-\tau)\theta^{\mu'}$ where $\tau \ll 1$).

The above procedure is repeated for a number of episodes until the actor learns an effective mapping from the system states to action values.
DDPG performs well in high dimensional spaces and for continuous input and output spaces.

\vspace{\medskipamount}\noindent\textit{Network details -- }
Our actor and critic networks consist of 3 dense layers with
ReLU activation units (actor: -300-300-30-; critic: -300-300-10-) except for the last actor layer that is $tanh$ to bound the action values. The learning rate was set to 0.001 and batch size to 64 for the ADAM optimizer in all experiments. The action bounds
were reduced from $[-1.0,1.0]$ to $[-0.2,0.2]$ during testing
for safety reasons.

\subsection{Monolithic controller}\label{sec:monolithic}
Our work focuses on multi-objective tasks where rewards for the
tasks can be dense or sparse. The object reaching task was
investigated by Pham~et~al. in \cite{Pham-2018-icra} and therefore
our monolithic controller is similar to their work. However,
to provide a more general solution we do not adopt their
constraint optimization layer which is task specific. Instead,
we sum all reward terms into a single composition reward
function: 
\begin{equation}\label{eq:reward}
    r = R_t + R_o + R_s,
\end{equation}
where $R_t$ is a dense reward for reaching the target, $R_o$ is a
sparse penalty for hitting an obstacle and $R_s$ is a sparse
reward for successfully reaching the target. It should be noted
that we do not need to give weights to the different rewards as
the penalty/reward values $p_o$ and $p_s$ implement this implicitly
(see Section~\ref{sec:experiments} for the ablation study).

$R_t$ is defined as the distance from the manipulator end-effector
(tooltip) to the target as $R_t = -d_t$. This reward term is continuous
and thus provides a dense reward for all states.
The distance penalty is defined in meters.

$R_o$ is defined as a sparse reward (penalty) if the distance of any
part of the robot arm and the obstacle, $d_o$, is less than a fixed threshold
$\tau_o$ as
\begin{equation}
R_o= 
\begin{cases}
    -p_o,  ~\text{if}~|d_o| < \tau_o \\
     0,  ~\text{otherwise}
\end{cases} \enspace .\\
\label{eq:Ro}
\end{equation}
$R_s$ is defined as a sparse positive reward for a successfully completed action (robot reaches the target) as
\begin{equation}
R_s= 
\begin{cases}
    p_s,   ~\text{if}~ d_t < \tau_s\\
    0,   ~\text{otherwise}
\end{cases}  \enspace , \\
\label{eq:Rs}
\end{equation}
which is active if no collision occurs during the same
time.

In the experiments, the parameters were set to the following
values based on preliminary experiments: $p_o = 4.0$, $p_s = 4.0$ and
$\tau_s$ = \SI{50.0}{\milli\meter}. In the experiments we report
various operation points by changing the value of
$\tau_o$ (Figure~\ref{fig:teaser}). Ablation study of the other
parameters is provided in the experiments as well.

\subsection{Hybrid controller}\label{sec:hybrid}
Our hybrid controller is essentially the same dual-mode controller as proposed by
Sangiovanni~\cite{Sangiovanni-2021-ieeecsl}. The original procedure is to train controllers for "obstacle avoiding" ($\mu_1$) and "goal reaching"
($\mu_2$) independently with DDPG and then combine them using a
"switching rule". Based on the original work the two reward functions
would be $r_1 = R_o$ and $r_2 = R_s$ (or $r_2=R_t+R_s$), but interestingly
we found that such hybrid controller is inferior to the variant
where reward functions are enforced to correlate by mixing terms
from each other.

The correlated hybrid controller is trained using the
following reward functions:
\begin{equation}
\begin{split}
    r_1 &= R_t + R_o + R_s ~\text{for}~\mu_1\\
    r_2 &= R_t + R_s ~\text{for}~\mu_2
\end{split} \enspace .
\end{equation}
In $r_1$ we emphasize the obstacle avoidance by setting
$p_o = 10.0$ and $p_s = 10.0$ and this controller effectively
avoids collisions but only rarely reaches the target (see
"DDPG avoidance" in Figure~\ref{fig:teaser}). On the other
hand, $r_2$ is solely trained for reaching the target
and thus it has high collision rate, but it almost always
reaches the target ("DDPG Reaching" in Figure~\ref{fig:teaser}).
Our approach to use common reward terms 
"glues" together the two policies and thus the "correlated" hybrid controller
is able to achieve low collision rate and high success rate beyond
the best result with the monolithic controller (Figure~\ref{fig:teaser}).

The final part of the hybrid controller is the
{\em switching rule}. We adopt the simple distance based
rule from Sangiovanni~et~al. as depicted in
Algorithm~\ref{alg:hybrid}. The hybrid controller operation
point is set by the switching threshold $\tau_{hyb}$.
By adjusting the threshold the robot can be
adjusted without re-training the controllers. The operation
points with various thresholds are shown in Figure~\ref{fig:teaser}.

\begin{algorithm}[h]
\SetAlgoLined

compute distance from the tool tip to the target $d_t$\;
compute distance from the robot to the obstacle $d_o$\;
   \If{$d_o < \tau_o$}{
   halt; \tcc{collision}
   }
   \If{$d_t<\tau_t$}{
   halt; \tcc{goal reached}
   }
  \eIf{$d_o < \tau_{hyb}$}{
   step $\mu_1$; \tcc{avoid}
   }{
   step $\mu_2$; \tcc{reach}
  }
\caption{Hybrid controller - one control step}
\label{alg:hybrid}
\end{algorithm}

\section{EXPERIMENTS}\label{sec:experiments}

\subsection{Set-up}
Both real and simulated experiments (see Figure~\ref{fig:teaser}) were performed using a Franka Emika Panda robotic arm with 7 degrees of freedom
and controlled by producing the joint velocities. The multi-objective task was to reach the stationary target while avoiding to hit the dynamic obstacle. In the simulated environment the target was placed randomly within a pre-defined  rectangular region
(shaded dark gray in Figure~\ref{fig:teaser}). The obstacle was moved randomly across the work space defined by a larger rectangle that
includes the target rectangle (light gray in Figure~\ref{fig:teaser}).
The same setup was implemented on a physical setup also depicted
in Figure~\ref{fig:teaser} where the white paper and table represent
the two rectangle regions.   

\vspace{\medskipamount}\noindent\textit{Simulator -- } CoppeliaSim~\cite{coppeliaSim} robot simulator was used to train the neural
controllers. The simulator was interfaced with the  PyRep API library~\cite{james2019pyrep}. The simulator computes the object
and robot joint coordinates. In the simulator, the distance $d_o$ between
the robot arm and obstacle was computed using the minimum Euclidean distance between the robot body and the obstacle surface. In
the physical setup the distance was approximated by computing
the minimum distance between each robot joint and the obstacle
center. Distance calculations and collision detection was automatically
provided by CoppeliaSim. The distance between the target and robot end-effector, $d_t$, was computed similarly in both the simulator
and the real setup using Euclidean distance between the target center
and the robot tooltip.

\vspace{\medskipamount}\noindent\textit{Real setup -- }
Franka Emika Panda robotic arm was interfaced through ROS
Robot Operating System~\cite{Quigley09}. ROS provides real-time
computation for the joint positions and velocities. The workspace was captured using an Intel RealSense D435 camera attached at the
top of the table. The obstacle and target were attached with
AprilTag markers~\cite{apriltag} for which the detection functions
provide 3D coordinates and orientation. The camera system was
calibrated to bring the marker coordinates to the common coordinates defined by the robot frame. The state distances were
computed as described above and the robot was controlled
using the ROS velocity control interface available for Panda.

\vspace{\medskipamount}\noindent\textit{State vectors -- }
The following state vectors were used in our experiments:
\begin{compactitem}
\item Monolithic: $s = \left(\gamma_{i},\dot{\gamma_{i}}, \vv{et}, \vv{eo}\right)$
\item DDPG avoidance: $s = (\gamma_{i},\dot{\gamma_{i}}, \vv{et}, \vv{eo})$
\item DDPG reaching: $s = (\gamma_{i},\dot{\gamma_{i}}, \vv{et})$,
\end{compactitem} 
where $\gamma_{i}$ and $\dot{\gamma_{i}}$ are the manipulator's intrinsic
position and velocity vectors that include a single value for each
joint. $\vv{et}$ is a directional scalar vector from the end-effector tool tip to the target center and $\vv{eo}$ is a directional scalar vector from end-effector to the obstacle center. The coordinate frame
origin is located at the base of the robotic arm.

\subsection{Simulation results}
The operation point of the monolithic controller in Section~\ref{sec:monolithic} is defined by the two
thresholds, $\tau_o$ for obstacle avoidance and
$\tau_s$ for reaching the target. If the thresholds
are changed the controller needs to be re-trained.
The operation point of the hybrid controller in Section~\ref{sec:hybrid}
is set by a single threshold, $\tau_{hyb}$, that defines the switching
point between the reaching and obstacle avoiding controllers,
$\mu_2$ and $\mu_1$, respectively.
The hybrid controller does not require re-training.

Both controllers were tested in the simulated environment
using a number of different threshold combinations
$\tau_o, \tau_{hyb} \in \{10, 20, 50, 100, 150, 200, 250\}$.
The three best results at the operation
points $\tau_o = \tau_{hyb}=\{50, 100, 150\}$
are shown in the graph in the first page (Figure~\ref{fig:teaser}). In all experiments $\tau_s = 50$.
These results clearly indicate that the monolithic controller cannot achieve the same success-collision rate trade-off as the hybrid controller. Moreover, as distinct advantage the hybrid controller
needs to be trained only once and therefore the optimal operation
point search is orders of magnitude faster. Interestingly,
the hybrid controller also achieved lower collision rate than
the collision avoidance only controller (cross) and almost the
same success rate as the reach-only controller (square) which
indicates that the hybrid controller rule of Sangiovanni et al.~\cite{Sangiovanni-2021-ieeecsl} has beneficial properties
for Sim-to-Real applications.

\begin{figure}[bh]
    \centering
    \includegraphics[width=0.9\linewidth]{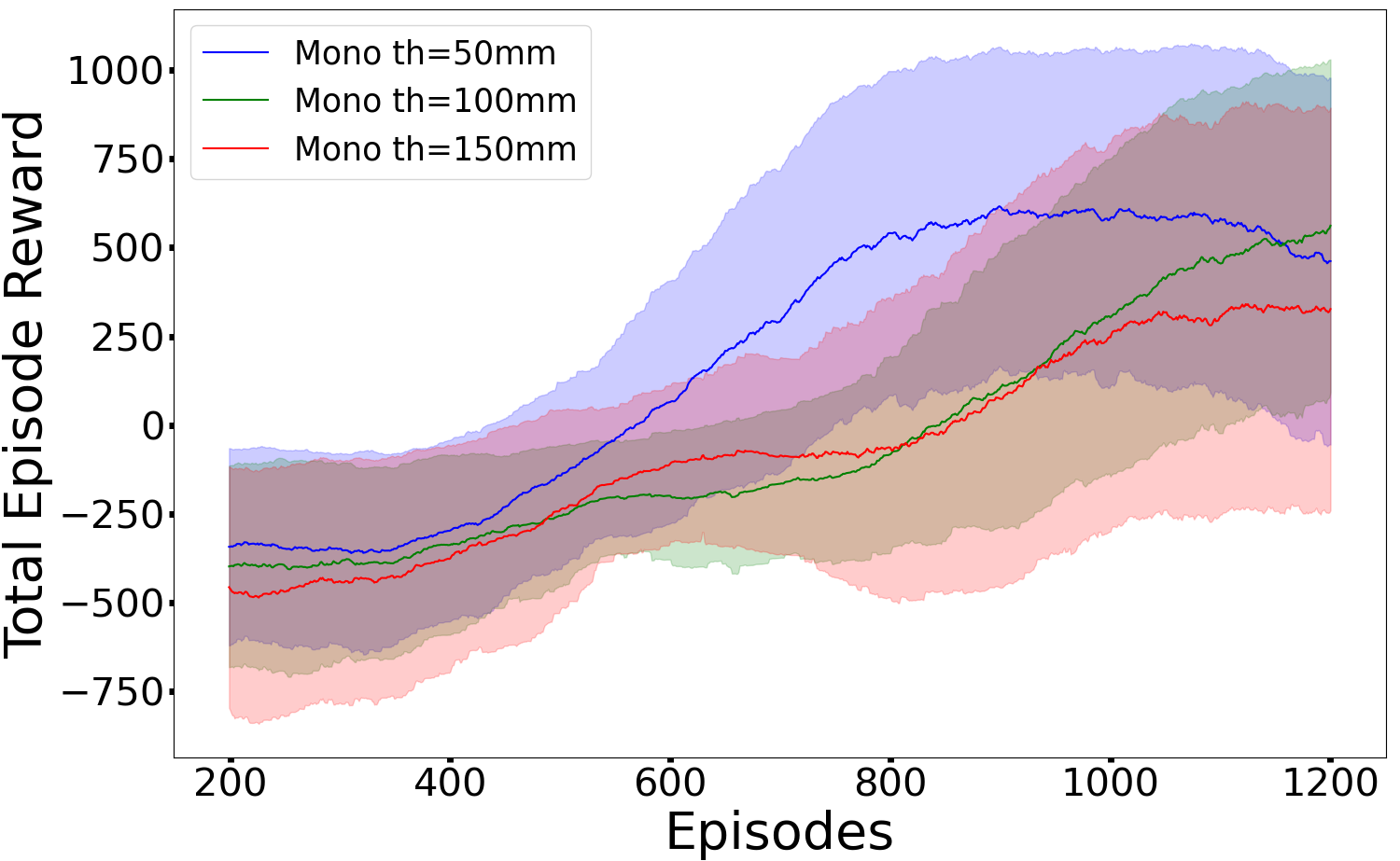}\\
    \includegraphics[width=0.9\linewidth]{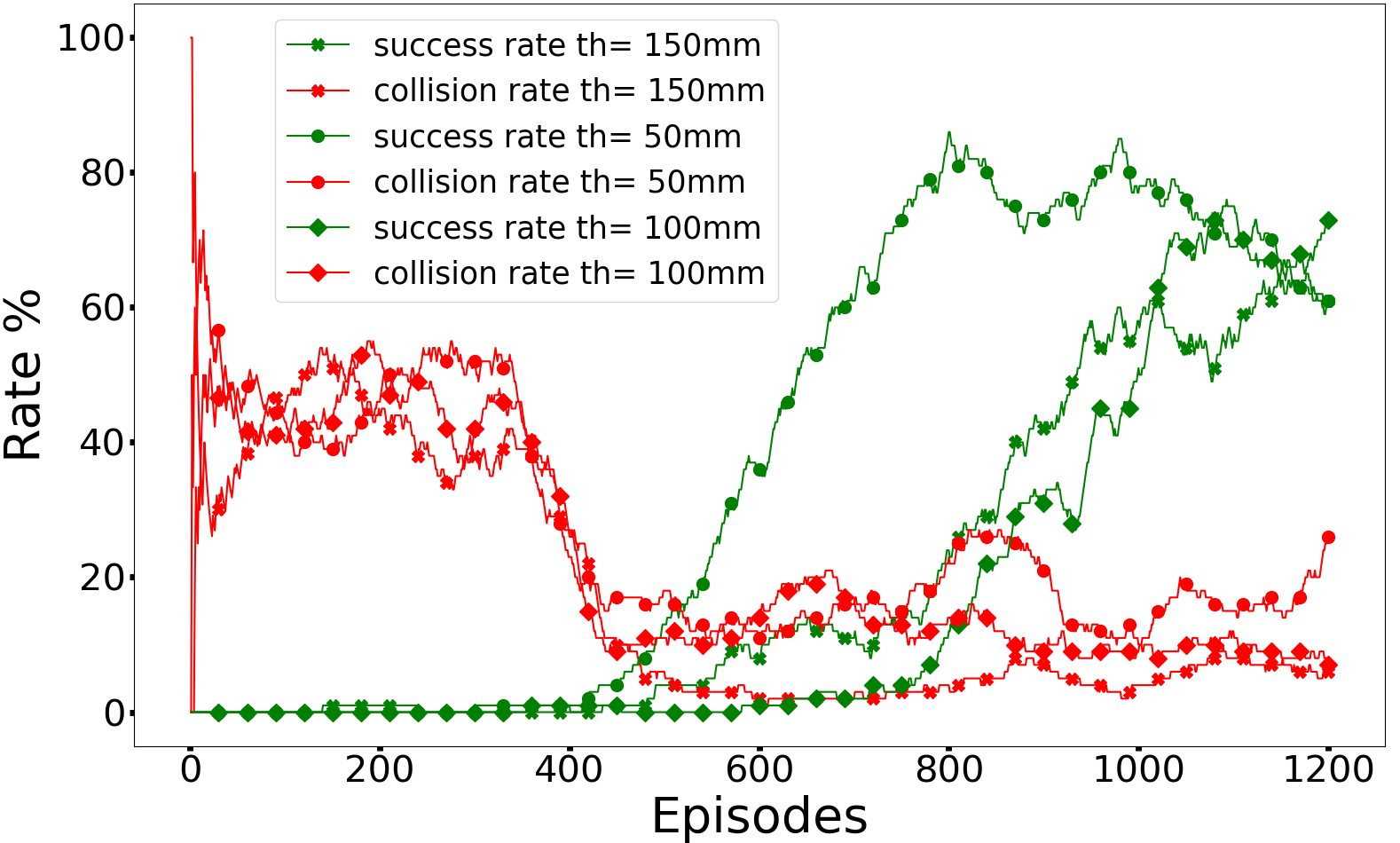}
    \caption{Convergence of the monolithic controller. Top: per episode rewards (running average as the solid line and true values as shaded regions) for different values of $\tau_o$. Bottom: Corresponding success and failure rates for different $\tau_o$.
    }
    \label{fig:exp_mono}
\end{figure}

\vspace{\medskipamount}\noindent\textit{Convergence properties -- }
Another important dimension of RL training is the number of simulated
episodes. To study the convergence properties we used the monolithic
controller as the study case as its training is more difficult than
training the single controllers for the hybrid controller. The per
episode rewards for various thresholds are shown in Figure~\ref{fig:exp_mono}. The average reward remains negative until approximately 500 episodes, meaning that in most of the episodes
the simulation ends to collision. Successful training requires approximately 1,000 episodes after which the average reward remains
clearly positive and thus the target is reached and collisions avoided
in most of the cases. This result can be verified from the success and failure rates in Figure~\ref{fig:exp_mono} (bottom) which remain
steady after 1,000 episodes. All controllers were trained 3 times and the best performing ones were chosen to reduce the affect of initial random guess. For all experiments, the monolithic controller training lasted 1200 episodes (approx. 2 hours), hybrid reaching was trained for 700 episodes (approx. 55 minutes), and hybrid avoiding - 1000 episodes (approx. 1 hour 50 min). We used GPU GeForce GTX 980M/PCIe/SSE2 and Intel® Core™ i7-6820HK CPU @ 2.70GHz x 8.

\begin{figure}[t]
    \centering
    \includegraphics[width=0.7\linewidth]{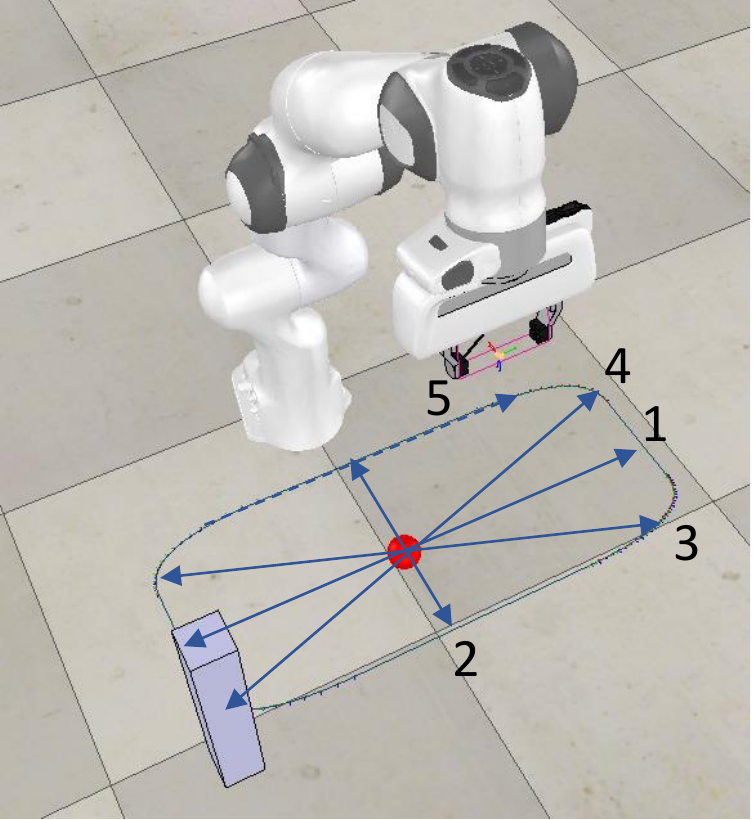}
     \caption{Five experimented scenarios to compare success and failure rates between the simulated and real experiments.
    \label{fig:scenarios}}
\end{figure}

\subsection{Sim-to-Real}

For the Sim-to-Real problem it is important that the performance
numbers obtained in the simulated environment are sufficiently
good estimates of the
expected performance in the real environment. For this purpose,
we trained the hybrid and monolithic controllers using the best
parameter settings ($\tau_o = 100~mm, \tau_s = 50~mm$ and $\tau_{hyb} =250~mm$)
and then computed the success and failure rates for five different
scenario that were easy to perform with the physical setup as well.
The scenarios are defined by the movement trajectory of the
obstacle for a fixed target position (Figure~\ref{fig:scenarios}).
Each experiment was repeated 10 times in the simulation and real environments using both controllers.
The obstacle velocity in the simulated environment was varied
between 0.018~m/s and 0.040~m/s to approximate the speed of
obstacle in real experiments where it was moved manually. 
The results from these experiments are summarized in Table~\ref{tab:scenarios}. Note that here we computed the distance $d_o$ in the same manner in simulator as in real set-up, i.e., as the minimum distance between each robot joint and the obstacle center. In testing, this distance is used only by the hybrid controller.

\begin{figure*}[h]
    \centering
    \subfloat[Scenario 2 - Monolithic cannot reach the target as the obstacle is too close (real: top, simulated: bottom)]
    {\includegraphics[width=1\linewidth]{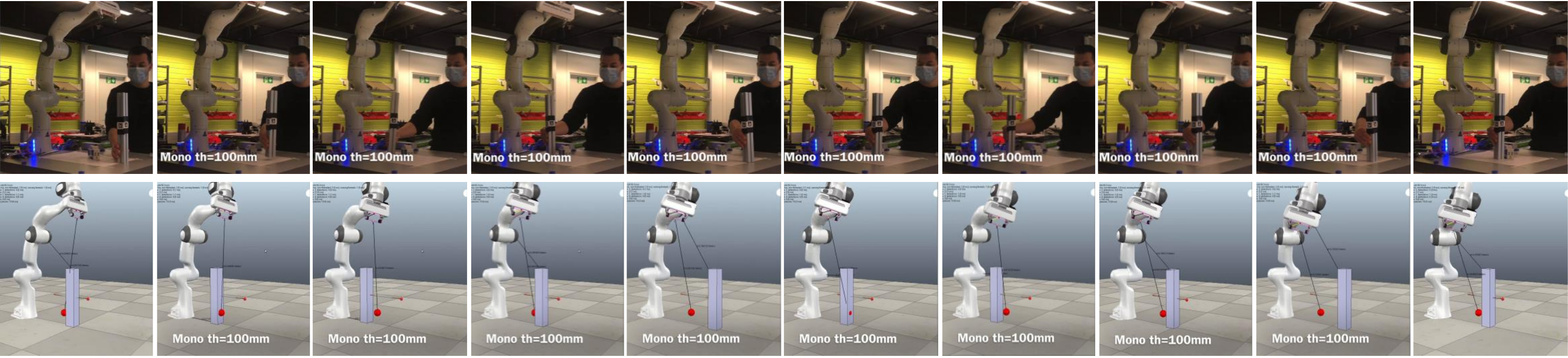}}\\
    \subfloat[Scenario 2 - Hybrid controller succeeds]
    {\includegraphics[width=1\linewidth]{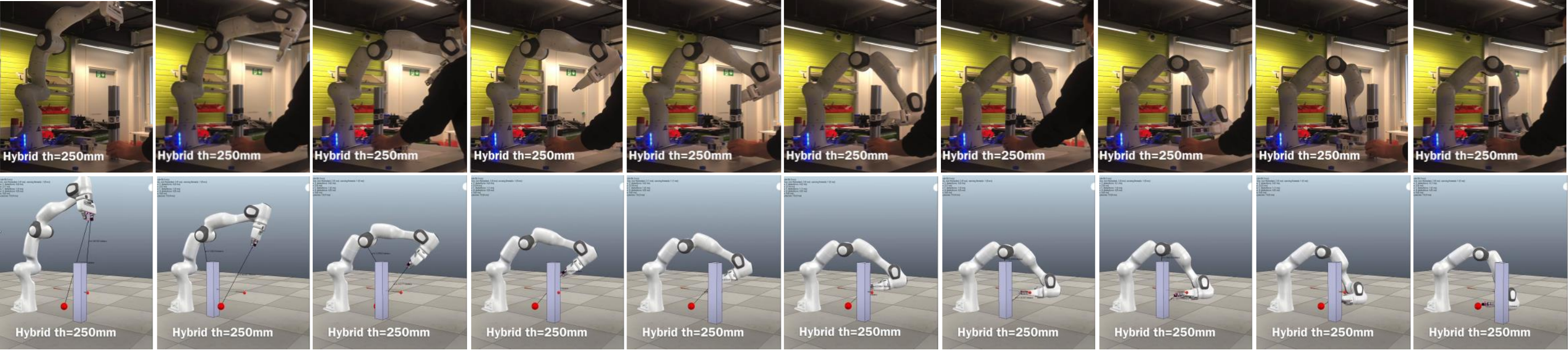}}\\
    \subfloat[Scenario 3 - Monolithic succeeds]
    {\includegraphics[width=1\linewidth]{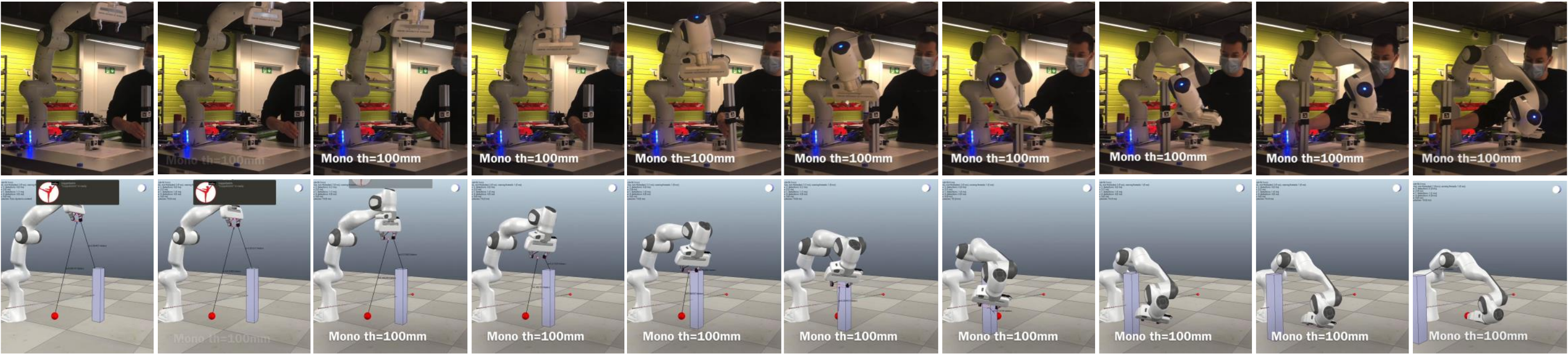}}\\
    \subfloat[Scenario 3 - Hybrid succeeds]
    {\includegraphics[width=1\linewidth]{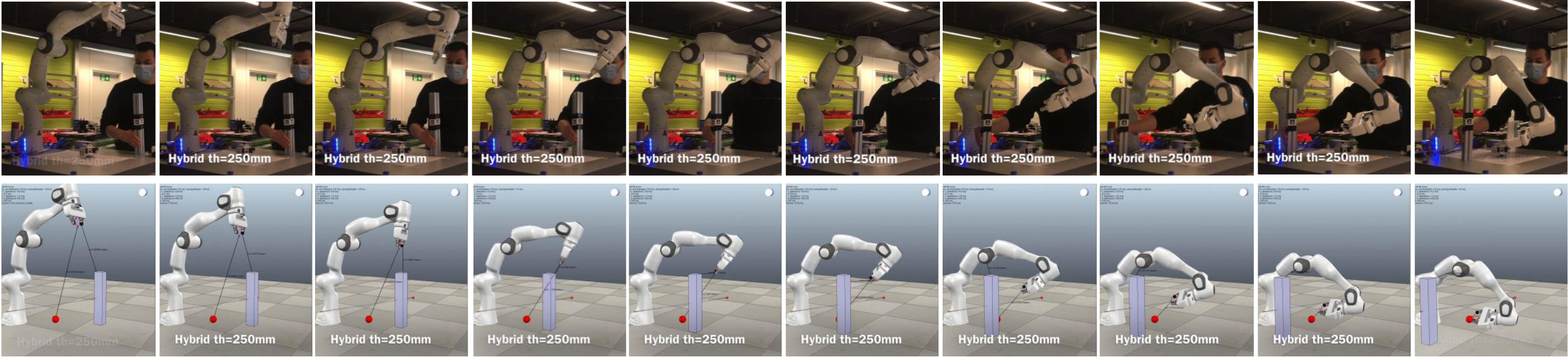}}
     \caption{Video frames from the Sim-to-Real experiments (\textcolor{blue}{See the supplementary material for the original videos}).}
    \label{fig:videos}
\end{figure*}

From the results two important findings can be made:
1) the hybrid controller systematically provides higher success
rates and lower failure rates than the monolithic controller
(from 100-30 success and 0-80 failure vs. 40-20 and 0-80) and
2) the real setup success and failure rates are within $\pm 20\%$
from those simulated even with the limited number of samples.
In Scenario~5 the monolithic controller
is slightly better in simulator, but obtains the same numbers
with the real setup.
Video frames from both simulated and real
episodes are shown in Figure~\ref{fig:videos}.

\begin{table}[t]\label{tab:scenarios}
\caption{The success and collision rates of the monolithic and hybrid controllers used in our experiments. The different scenarios correspond to the five different obstacle trajectories. Note that all controllers were trained in the simulator using random obstacle trajectories. In Scenario 2 the monolithic controller remains stationary (Figure~\ref{fig:videos}).}
\resizebox{1.0\linewidth}{!}{
\begin{tabular}{l rr rr}
\toprule
 & \multicolumn{4}{c}{\textit{Success/collision rate [\%]}}\\ & \multicolumn{2}{c}{Hybrid}  & \multicolumn{2}{c}{Mono} \\

& sim & real & sim & real  \\ \hline
Scenario 1 & 100 / ~0 & {\bf 90 / 10} & 40 / 60 & 30 / 70  \\
Scenario 2 & 30 / 40 & {\bf 30} / 40 & ~0 / ~0 & ~0 / {\bf ~0}   \\
Scenario 3 & 100 / ~0 & {\bf 80 / 20} & 20 / 80 & 30 / 70  \\
Scenario 4 & 80 / 20 & {\bf 60 / 40} & 40 / 60 & 30 / 70   \\
Scenario 5 & 20 / 80 & {\bf 30 / 70} & 40 / 60 & {\bf 30 / 70}  \\

\bottomrule
\end{tabular}\label{tab:scenarios}
}
\end{table}

\begin{table}[t]
\caption{Success and collision rates for different combinations of $p_o$ and $p_s$ with other parameters fixed, average for 1000 test-runs in simulator. The highest success and lowest failure rates are achieved by the two extreme ends of the penalty combinations as it is expected. The marked value ($p_o = p_s = 4.0$ was used in all other experiments due to its good success/failure rate trade-off.}\label{tab:ablation}
\resizebox{1\linewidth}{!}{
\begin{tabular}{l c r r r r}
\toprule
& \multicolumn{4}{c}{$p_s$}\\
& & 1.0 & 2.0 & 4.0 & 8.0 \\
\midrule
\multirow{4}{*}{$p_o$}
& 1.0  & 95.7 / ~8.6 & 97.7 / ~6.6 & 95.3 / ~9.2 & {\bf 99.0} / ~9.1 \\
& 2.0  & 74.5 / ~2.1 & 86.7 / ~4.4 & 89.4 / ~4.7 & 91.2 / ~7.7\\
& 4.0  & 11.5 / ~3.5 & 45.4 / ~2.2 & \fbox{87.4 / 3.4}    & 87.4 / ~5.7             \\
& 8.0  & 11.6 / {\bf ~0.7} & 15.7 / ~2.0 & 23.1 / ~5.3 & 40.5 / ~0.9             \\
\bottomrule
\end{tabular}}
\end{table}

\subsection{Ablation study}
The performance of the monolithic controller is heavily affected
by the magnitude of the penalties $p_o$ of the reward term
$R_o$ in (\ref{eq:Ro}) and $p_s$ of $R_s$ in (\ref{eq:Rs}). These
define how much reward (negative penalty) is given when the
object is reached and how much penalty is given for nearly colliding
with the obstacle. To make sure that the selected values $p_o = p_s = 4.0$ are optimal, we conducted an ablation study where various values
were used and average success and failure rates were computed over 1,000
simulated test episodes. The results are summarized in Table~\ref{tab:ablation}.

The values used in all previous experiments are
$p_o = 4$ and $p_s = 4$ clearly provide a good trade-off
between the success and collision (failure) rates. In addition, the monolithic
controller performance is more sensitive to the obstacle penalty $p_o$ as
for values greater than 4.0 the success rate drops significantly. 4.0 is the limit
after which failure rate is still tolerable and the success rate can be
adjusted using the success penalty (reward) $p_s$. 

\section{CONCLUSION}\label{sec:conclusion}
We implemented and compared two controller architectures
for the multi-object Sim-to-Real DDPG reinforcement
learning: monolithic and hybrid. The most important finding is that
the hybrid architecture by Sangiovanni et al.~\cite{Sangiovanni-2021-ieeecsl}
provides clearly better success-failure trade-off and is easier to
train as the DDPG needs to be conducted only once for the
single task controllers. The future research shall focus on learning the hybrid switching rule automatically, extending the range of tested scenarios, and exploring the hybrid/monolithic approaches for a wider range of multi-objective tasks.

\bibliographystyle{ieeetr}
\bibliography{sim2real}

\end{document}